\documentclass{article} 
\usepackage{iclr2027_conference,times}

\usepackage{amsmath,amsfonts,bm}

\def\eqref#1{equation~\ref{#1}}

\def\1{\bm{1}}

\DeclareMathAlphabet{\mathsfit}{\encodingdefault}{\sfdefault}{m}{sl}
\SetMathAlphabet{\mathsfit}{bold}{\encodingdefault}{\sfdefault}{bx}{n}

\usepackage{hyperref}
\usepackage{url}
\usepackage{graphicx}
\usepackage{booktabs}
\usepackage{xcolor}
\usepackage{colortbl}
\definecolor{gen1}{RGB}{243,246,249}
\definecolor{gen2}{RGB}{228,237,244}
\definecolor{gen3}{RGB}{196,214,228}
\definecolor{tmp1}{RGB}{252,246,236}
\definecolor{tmp2}{RGB}{245,230,206}
\definecolor{tmp3}{RGB}{232,208,172}
\definecolor{lng1}{RGB}{241,246,237}
\definecolor{lng2}{RGB}{220,232,212}
\definecolor{lng3}{RGB}{190,214,178}
\definecolor{gainbadge}{RGB}{198,122,48}
\usepackage{amssymb}
\usepackage{float}

\title{RESUME: Recurrent State Updates from Motion and Residual Signals for Efficient Video Language Modeling}

\author{\textbf{Can Zhang\textsuperscript{1,2}, Xiaotian Han\textsuperscript{2}\thanks{Project Lead.}, Junyuan Shang\textsuperscript{2}, Yuchen Ding\textsuperscript{2}, Zhenyu Zhang\textsuperscript{2},}\\
\textbf{Shuohuan Wang\textsuperscript{2}, Dianhai Yu\textsuperscript{2}, Ruirui Li\textsuperscript{1}\thanks{Corresponding author.}}\\[4pt]
\textsuperscript{1}Beijing University of Chemical Technology\qquad\textsuperscript{2}Baidu, Inc.\\[2pt]
\texttt{\small alexlessend@gmail.com, \{hanxiaotian, shangjunyuan, dingyuchen,}\\
\texttt{\small zhangzhenyu07, wangshuohuan, yudianhai\}@baidu.com, ilydouble@gmail.com}}

\newcommand{\dbadge}[1]{\textsubscript{\textcolor{gainbadge}{\bfseries+#1}}}

\iclrfinalcopy 
\begin{document}

\maketitle
\lhead{} 

\begin{abstract}
Existing video language models encode sampled RGB frames independently, so a long video must either exhaust the token budget or drop the changes between sampled frames. Codec-aware front-ends read the motion vectors and residuals that encoding already produced, but in their deployed form each predictive frame is still tokenized on its own: the tokens are a function of the current primitives, not of a carried reference. We argue that a more natural function is of both---the current primitives and a carried reference. A clip and its time reversal share the same frames and differ only in the order of changes---an axis that symmetric pooling discards by construction, and that is non-empty in the frozen vision features VideoLMs actually use---and the codec recurrence already composes those changes in order against a reference state. We introduce RESUME, a stateful codec representation: an anchor I-frame initializes a compact latent state, each subsequent predictive frame is consumed as an update to that state, and a shared readout exposes VideoLM-compatible tokens from the accumulated state. Codec prediction is thereby kept at the representation level and handed to the language model as a trajectory, not as a set of independent token groups. At the same per-predictive-frame token budget as prior codec-aware methods, a predictive frame enters the language model as a readout of what the front-end already knows, not as an encoding of the current primitives alone. Across ten benchmarks, the gains concentrate on temporal reasoning: on all three temporal benchmarks RESUME improves over both the RGB-frame baseline LLaVA-Video-7B (by $2.8$, $5.1$, and $3.9$ points on TempCompass, TOMATO, and MVBench) and the codec-based baseline CoPE-7B, while staying competitive on general and long-form QA. Frozen-transition tests further show anchor dependence, order sensitivity, and useful rollout behavior beyond the training horizon.
\end{abstract}

\section{Introduction}
\label{sec:intro}

Video language models encode visual observations so that a language model can reason about actions, event order, and cross-frame relations. Recent systems improve video question answering through stronger vision backbones, larger instruction corpora, and longer context windows~\citep{damonlpsg2023videollama,zhang2024videoinstructiontuningsynthetic}. Their standard abstraction, however, remains a collection of independently encoded RGB snapshots.

Dense RGB sampling preserves transient actions and state changes but repeatedly pays the full image-encoding and token cost. Sparse keyframe sampling is cheaper but discards the changes between sampled frames; the central limitation is therefore representational rather than a lack of context length.

The question is representational: what must a video representation preserve for temporal questions to be answerable? A clip and its time reversal contain the same frames but differ in the order of changes, so symmetric pooling is invariant to reversal by construction. A training-free probe study on frozen vision encoders confirms that static pooling is exactly invariant, while an order-sensitive probe flips on $75$--$86\%$ of reversed clips (Figure~\ref{fig:hierarchy}). Temporal order is therefore a distinct information axis discarded by symmetric aggregation and preserved by a state updated in sequence.

Video codecs already exploit temporal redundancy by storing an independently coded I-frame and describing each predictive frame with motion vectors and residuals~\citep{mpeg,Wiegand2003OverviewOT}. Compressed-domain recognition established that these primitives carry usable dynamics~\citep{wu2018coviar,teamnet,biswas2025scalablemodelingcompressedvideos}, and codec-aware VideoLMs such as Video-LaVIT~\citep{videolavit2024}, EMA~\citep{ema2025}, and CoPE-VideoLM~\citep{cope2026} use them to cover more timestamps at lower visual-token cost. However, their deployed encoders emit an independent token group for each predictive frame, leaving reference dependence and temporal composition to language-model attention.

The codec recurrence states more than that $\tau_t$ and $\delta_t$ are sparse:
\begin{equation}
\label{eq:codec-recurrence}
\hat I_t = \mathrm{Warp}(\hat I_{t-1}, \tau_t) + \delta_t.
\end{equation}
A predictive frame is a \emph{function of a reference state}. Three consequences follow. First, \textbf{reference dependence}: the same rightward motion field means ``the ball moves right'' against one reference and ``the person shifts right'' against another, so $(\tau_t,\delta_t)$ alone does not determine the updated content. Second, \textbf{compositionality}: video events are typically accumulations of small changes, which become an event only when applied in sequence to a state that keeps changing. Third, \textbf{path dependence}: the outcome depends on the order in which updates are applied, not on the set of updates. These are not properties imposed on the data; they are what the recurrence already computes. They are preserved by construction in a state that is updated in sequence, and left to be reconstructed by language-model attention when each predictive frame is tokenized independently. Prior codec-aware work is not unaware of the recurrence: CoPE-VideoLM emulates one warping step in feature space during pre-training. But that step reads its reference from a decoded RGB frame rather than from a carried state, and it is removed before the encoder meets the language model, so nothing accumulates and nothing survives into inference.

The question is therefore not how to represent each predictive frame with fewer tokens, but how a VideoLM front-end can keep the codec's own predictive structure at the representation level, from one anchor to the next, rather than leaving it for language-model attention to reconstruct.

\begin{figure}[t]
\centering
\includegraphics[width=\textwidth]{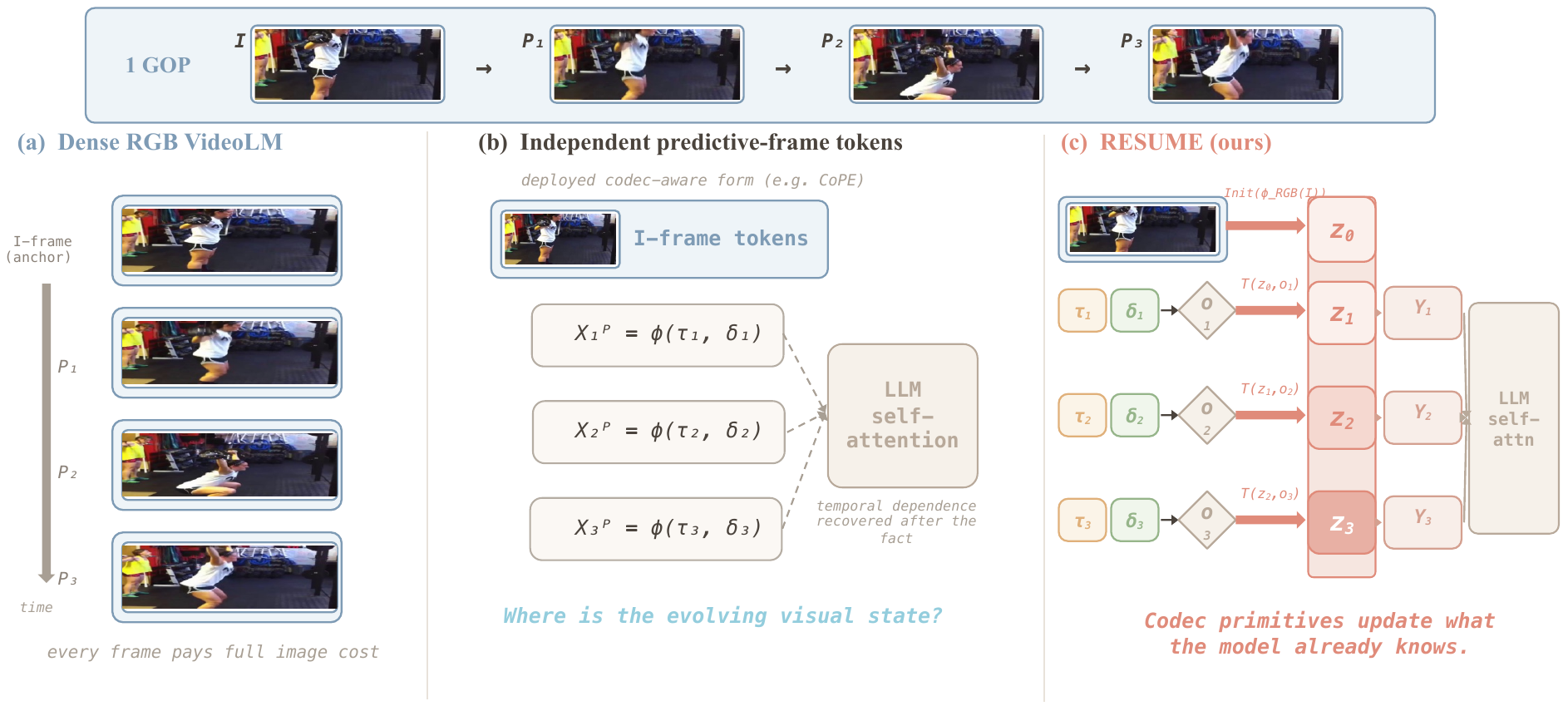}
\caption{Three ways to turn a group of pictures into visual tokens.
(a)~Dense RGB encoding treats every frame as a full image.
(b)~Deployed codec-aware tokenization reads motion vectors and residuals, but emits an independent token group per predictive frame; temporal dependence is left to language-model attention.
(c)~RESUME initializes a compact latent state from the anchor I-frame and updates that same state with each codec observation; the tokens of a predictive frame are a readout of the accumulated state.
The two codec columns tokenize each predictive frame; only (c) maintains a reference-dependent state in the visual front-end.}
\label{fig:teaser}
\end{figure}

We introduce \textbf{RESUME}, a stateful codec representation that carries a latent state across an anchor's predictive frames instead of tokenizing each one independently (Figure~\ref{fig:teaser}). Each anchor I-frame initializes a compact latent state, subsequent motion-residual observations update that state causally, and a shared readout exposes VideoLM-compatible tokens from the accumulated state. The state is reset at the next anchor, while frozen vision features provide its initial condition and supervision space. Unlike independent tokenization, RESUME learns \texttt{previous state + current primitive $\to$ next state} at the same per-predictive-frame token budget.

Training has two stages. Stage 1 learns the transition from multi-step ordered rollouts, supervising both intermediate readouts and their changes. Stage 2 attaches the frozen transition to a VideoLM, packs the state readouts with each I-frame token group, and trains the language model to use them through a dedicated projector.

The experiments support both the representation-level claim and its downstream effect. In a training-free probe study, the order-sensitive probe flips on $75$--$86\%$ of reversed clips, while static pooling is exactly invariant and order-agnostic difference statistics move only through endpoint displacement. On the frozen transition, applying the same updates to different anchors preserves anchor-dependent context, the correct update order wins on $65\%$ of videos, and the carried state is more accurate than a memoryless control on $75\%$ and $68\%$ of videos at rollout horizons of $8$ and $16$ steps. With the language model, RESUME improves over LLaVA-Video-7B by $2.8$, $5.1$, and $3.9$ points on TempCompass, TOMATO, and MVBench, and improves over CoPE-7B by $0.5$, $1.7$, and $0.6$ points, respectively, while using the same per-predictive-frame readout budget. These results connect the information hierarchy to the final VideoLM interface: preserving ordered codec changes in a carried state improves the benchmarks that require temporal evolution.

Our contributions are:
\begin{enumerate}
\item \textbf{Problem formulation.} We formulate codec-aware video representation as a causal latent state transition between successive anchors. This formulation identifies reference dependence, compositionality, and path dependence as the properties that independent predictive-frame tokenization leaves to language-model attention to reconstruct.
\item \textbf{Stateful codec representation.} We instantiate this formulation with RESUME: an anchor I-frame initializes a compact latent state, motion vectors and residuals are fused into one codec observation, and a causal transition carries the state across predictive frames. The resulting state encoder contains approximately $21$M parameters, while a shared readout exposes VideoLM-compatible tokens at the same per-predictive-frame budget as prior codec-aware methods.
\item \textbf{Empirical evidence.} Training-free probes on frozen vision encoders show that temporal order is a non-empty information axis discarded by symmetric aggregation (Figure~\ref{fig:hierarchy}). Frozen-transition experiments verify anchor dependence, order sensitivity, and useful rollout behavior beyond the training horizon (Figure~\ref{fig:mech}). On all three temporal benchmarks, RESUME improves over both the RGB-frame baseline LLaVA-Video-7B (by $2.8$, $5.1$, and $3.9$ points) and the codec-based baseline CoPE-7B, while staying competitive on general and long-form QA.
\end{enumerate}

\section{Related Work}
\label{sec:related}

\subsection{Frame-based VideoLMs and post-hoc compression}
\label{sec:rw-frame}

VideoLMs extend image VLMs by supplying visual tokens at multiple timestamps. Video-LLaMA~\citep{damonlpsg2023videollama}, VideoChat2~\citep{2023videochat}, VideoLLaMA2~\citep{damonlpsg2024videollama2}, LLaVA-NeXT-Video~\citep{liu2024llavanext}, and LLaVA-Video~\citep{zhang2024videoinstructiontuningsynthetic} established the standard workflow: sample RGB frames, tokenize each independently, and let a temporal module or the language model reason over the sequence. Subsequent work reduces the resulting token count by pooling, merging, pruning, or learnable resampling~\citep{liblip2,yao2024minicpm,bolya2022tome,fastv,visionzip,dycoke,llavascissor}. These compressions are post-hoc: they operate after a dense RGB representation has been produced, so the encoding cost is already paid and each retained frame remains an independent observation.

\subsection{Codec primitives as VideoLM input}
\label{sec:rw-codec}

Compressed-domain recognition showed that I-frames, motion vectors, and residuals carry usable dynamics without decoding most frames~\citep{wu2018coviar,teamnet,biswas2025scalablemodelingcompressedvideos}. Video-LaVIT~\citep{videolavit2024} discretizes motion into language-like tokens but largely discards residuals; EMA~\citep{ema2025} aggregates an I-frame and motion into a fixed-length GOP summary, collapsing P-frame order. A further group uses codec signals as a \emph{selection} cue---keeping salient patches or refreshing a cache according to bitstream activity---which reduces what is encoded without changing what a predictive frame's representation is a function of~\citep{magevl2026,onevisionencoder2026,llavaonevision2_2026}. AdaCodec~\citep{adacodec2026} regresses codec-derived tokens onto a frozen teacher and drops the auxiliary head before language-model training.

CoPE-VideoLM~\citep{cope2026} encodes motion vectors and residuals into compact per-predictive-frame tokens. Its pre-training emulates a feature-space warping step using a reference re-measured from decoded RGB, but the auxiliary modules are removed before VideoLM integration. At inference, each predictive-frame token group depends on the current primitives rather than a carried visual state; temporal order is retained in the token sequence.

Codec-driven state propagation also appears outside this interface. Compressed-domain recognition maintains separate motion and residual states without I-frame initialization~\citep{biswas2025scalablemodelingcompressedvideos}, while video super-resolution propagates hidden states for reconstruction rather than language-model readout. ReMoRa~\citep{remora2026} uses a bidirectional I-frame--motion scan with one GOP-level readout and no residual input. RESUME instead combines causal GOP-local updates, I-frame initialization, early motion-residual fusion, and per-predictive-frame readouts at the same token budget as CoPE.

\section{Method}
\label{sec:method}

\begin{figure}[t]
\centering
\includegraphics[width=\textwidth]{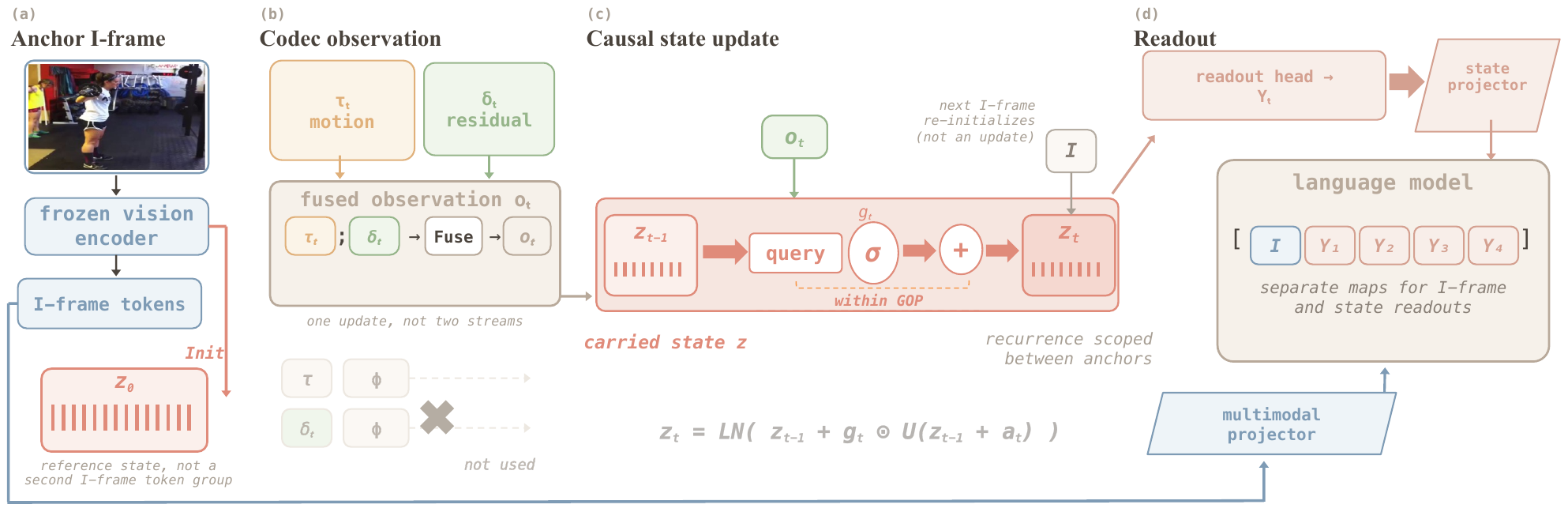}
\caption{RESUME as a codec-driven transition system.
An anchor I-frame initializes a compact latent state $z_0$ and, in parallel, keeps its own token path into the language model. Each predictive frame contributes one fused observation $o_t$ from motion vectors and residuals, written into the carried state. A shared head reads $Y_t$ from $z_t$ at the same per-predictive-frame token budget used by prior codec-aware front-ends. The state is reset at the next I-frame. A dedicated projector adapts state readouts to the language embedding space; the original multimodal projector is left untouched for I-frame tokens.}
\label{fig:arch}
\end{figure}

The question above is how a VideoLM front-end can keep the codec's own predictive structure at the representation level, rather than leaving it for language-model attention to reconstruct. We instantiate that as a transition system (Figure~\ref{fig:arch}). The architecture below is one realization; the constraints come from the three properties the codec recurrence already computes: reference dependence, compositionality, and path dependence.

\subsection{From independent tokenization to state transition}
\label{sec:method-transition}

A video $V=(F_1,\dots,F_T)$ is organized into groups of pictures (GOPs). Each GOP begins with an intra-coded I-frame and continues with predictive frames. Following the codec convention we write
\begin{equation}
F_t = \begin{cases}
I_t, & F_t \text{ is intra-coded},\\
P_t = (\tau_t, \delta_t), & F_t \text{ is predictive},
\end{cases}
\end{equation}
with block-wise motion vectors $\tau_t \in \mathbb{R}^{H_G \times W_G \times 2}$ and residuals $\delta_t \in \mathbb{R}^{H \times W \times 3}$. We use only I- and P-frames: bidirectional B-frames require future references and break the causal order that both streaming and autoregressive language modeling assume.

Independent predictive-frame tokenization implements $X^P_t = \phi(\tau_t, \delta_t)$: the tokens of a predictive frame are a function of the current primitives alone. The codec recurrence~\eqref{eq:codec-recurrence} is not of this form. It is a transition---the next frame is a function of a reference and an update---so the front-end we implement is a transition system,
\begin{equation}
\label{eq:resume}
z_0 = \mathrm{Init}\big(\phi_{\mathrm{RGB}}(I_0)\big), \qquad
o_t = \mathrm{Obs}(\tau_t, \delta_t), \qquad
z_t = \mathcal{T}(z_{t-1}, o_t), \qquad
Y_t = \mathrm{Read}(z_t),
\end{equation}
where $z_t$ is a persistent latent state, $o_t$ is the codec observation at step $t$, and $Y_t$ is an on-demand readout in the vision-encoder feature space.

The binding is what makes the three properties design constraints: $z_0$ is initialized from the codec's own anchor, $o_t$ is the codec-native pair $(\tau_t,\delta_t)$, $\mathcal{T}$ is causal and is the representation that reaches the language model, and $Y_t$ is read out at the same per-predictive-frame budget as prior codec-aware work. Reference dependence requires $\mathcal{T}$ to take $z_{t-1}$ as an argument. Compositionality requires updates to accumulate in one latent variable. Path dependence requires $\mathcal{T}$ not to be permutation-invariant in $t$. The rest of this section is one realization of these constraints, and the training that makes them the solution the model actually finds.

\subsection{Realizing the transition}
\label{sec:method-encoder}

\paragraph{Anchor as initial condition.}
Given the I-frame of a GOP, a frozen vision encoder produces patch tokens $X_I$. A state initializer projects them to the state width, concatenates a compact set of learnable queries $q$, and keeps the query positions after a stack of transformer layers:
\begin{equation}
z_0 = \mathrm{LN}\Big(\mathrm{Layers}\big([\,q;\, W_{\mathrm{in}} X_I\,]\big)_{q}\Big).
\end{equation}
This $z_0$ is not a second token group for the I-frame---the I-frame keeps its own path into the language model---but the reference on which every subsequent codec update operates. Without it, $\mathcal{T}$ would have nothing to condition on, and reference dependence would be vacated.

\paragraph{One observation per codec update.}
Motion states how existing content should be displaced; the residual states what motion compensation cannot explain. They are two halves of one codec step, not two independently consumable modalities. Encoding them in separate branches and concatenating the results, as prior codec-aware encoders do, would reintroduce the decomposition that the recurrence forbids. We therefore fuse them at a common spatial position. Motion vectors are patchified and embedded; residuals are embedded by a convolutional stem whose stride matches the motion grid; the two streams are concatenated position-wise and fused to state width:
\begin{equation}
o_t = \mathrm{Fuse}\big([\,\mathrm{MotionEmb}(\tau_t)\,;\,\mathrm{ResEmb}(\delta_t)\,]\big).
\end{equation}
The observation that enters $\mathcal{T}$ is therefore one innovation, matching the right-hand side of the codec recurrence.

\paragraph{Causal state update.}
At each step the state slots query this observation and write a gated residual into the carried state:
\begin{equation}
a_t = \mathrm{CrossAttn}\big(\mathrm{LN}(z_{t-1}),\, \mathrm{LN}(o_t)\big), \quad
g_t = \sigma\big(W_g[\,z_{t-1};a_t\,]\big), \quad
z_t = \mathrm{LN}\big(z_{t-1} + g_t \odot U(z_{t-1}+a_t)\big),
\end{equation}
with $U$ a small MLP. Cross-attention lets each slot select the spatial part of the update that concerns the content it holds; the gate controls how much of the candidate is written, the latent analogue of a residual being applied only where motion compensation left error. This is one realization of $\mathcal{T}$: it takes $z_{t-1}$ as an argument, accumulates in one variable, and is ordered in $t$. Other operators with the same interface would serve the same principle.

\paragraph{Readout of the accumulated state.}
The tokens that reach the language model are not an encoding of $(\tau_t,\delta_t)$. They are a readout of $z_t$. A compact set of learnable queries $r$ attends over the state and is projected back to the vision-encoder width:
\begin{equation}
Y_t = W_{\mathrm{out}}\,\mathrm{LN}\Big(\mathrm{Layers}\big([\,r;\, z_t\,]\big)_{r}\Big).
\end{equation}
Because the same head is applied at every step, readouts at different timestamps live in one coordinate system, and the language model receives a trajectory rather than a set of independent token groups. Readout frequency is then a free variable at inference---dense for fine-grained motion, sparse for long videos---without changing $\mathcal{T}$.

\subsection{Learning a trajectory, then reading it}
\label{sec:method-train}

The transition makes the three properties possible; training has to make them the solution the model finds.

\paragraph{Stage 1}
learns the transition in isolation. From an anchor I-frame the model consumes a short sequence of temporally ordered updates,
\begin{equation}
z_0 \xrightarrow{\;o_1\;} z_1 \xrightarrow{\;o_2\;} \cdots \xrightarrow{\;o_T\;} z_T,
\end{equation}
and the shared head reads out every intermediate state, not only the last. Targets come from the same frozen vision encoder applied to the true target frames, pooled so that a readout and its target are comparable:
\begin{equation}
\bar X_t = \mathrm{Pool}\big(\phi_{\mathrm{RGB}}(I_t)\big).
\end{equation}
The objective has three terms:
\begin{equation}
\label{eq:stage1-loss}
\mathcal{L} =
\underbrace{\frac{1}{T}\sum_t \|Y_t - \bar X_t\|_2^2}_{\text{state matching}}
\;+\;
\lambda_{\cos}
\underbrace{\frac{1}{T}\sum_t \big(1-\cos(Y_t, \bar X_t)\big)}_{\text{direction}}
\;+\;
\lambda_{\Delta}
\underbrace{\frac{1}{T-1}\sum_t \big\|(Y_t - Y_{t-1}) - (\bar X_t - \bar X_{t-1})\big\|_2^2}_{\text{trajectory}}.
\end{equation}
The first two terms fit each readout to a semantic endpoint. The third fits the \emph{change} between consecutive readouts, without which a model can score well by mapping each state to a plausible frame embedding while leaving the trajectory unconstrained.

\paragraph{Stage 2}
attaches the pre-trained transition to a released VideoLM without changing the language architecture or its objective. For each selected GOP, the I-frame initializes $z_0$ and the GOP's motion-residual updates drive $\mathcal{T}$. The shared head reads the intermediate states; the readouts are packed in temporal order and interleaved with the I-frame tokens for the language model. A predictive frame therefore enters the language model as a compact readout of accumulated state, not as a full image.

\subsection{Predictions and scope}
\label{sec:method-claims}

The formulation makes predictions that isolate carrying state. Keeping the base VideoLM's sampled frames fixed, and interleaving readouts of the predictive frames between them, should improve temporal, motion, and state-change reasoning over that sampling alone. Applying a fixed $(\tau,\delta)$ to different anchors should change the readout in a content-consistent way (reference dependence); shuffling updates between two anchors should degrade matching, whereas resetting the state before every update is invariant to that perturbation (path dependence). A model trained on short rollouts should extrapolate to longer horizons with error above a per-step re-initialization oracle and below a memoryless control. That curve diagnoses extrapolation of the transition; it is not evidence of a state maintained across a whole video, because inference re-initializes at every anchor.

Four boundaries follow. Stage-1 supervision comes from a frozen vision encoder: semantic features define the initial condition and the measurement space, while codec primitives parameterize the transition. The latent state carries no physical annotation, so we call it a codec-conditioned latent state. Efficiency is a property of the codec domain; the claim is the representational structure obtained at the same per-predictive-frame cost. And the recurrence spans the predictive frames between two anchors, not a whole video: an I-frame re-initializes the state. Claims concern the representation of a group of pictures; long-range aggregation remains the language model's responsibility.

\section{Experiments}
\label{sec:exp}

We first report question-answering results across general, temporal, and long-form benchmarks, then examine the information hierarchy and test whether the transition carries state across predictive updates.

\subsection{Evaluation benchmarks}
\label{sec:exp-bench}

We evaluate the model on ten benchmarks spanning general video question answering, temporal and motion reasoning, and long-form video. General video question answering: Video-MME~\citep{fu2024videomme}, PerceptionTest~\citep{patraucean2023perception}, NExT-QA~\citep{xiao2021next}, and ActivityNet-QA~\citep{yu2019activityqa}. Temporal and motion reasoning: TempCompass~\citep{liu2024tempcompass}, TOMATO~\citep{shangguan2024tomato}, and MVBench~\citep{li2024mvbench}. Long-form video: LVBench~\citep{wang2024lvbench}, Video-TT~\citep{zhang2025videott}, and Video-MMMU~\citep{hu2025videommmu}. Each video uses at most $64$ I-frames, and each of the four predictive updates per I-frame produces one $8$-token readout. ActivityNet-QA is scored by Claude Opus~4.8~\citep{anthropic2026claudeopus48}. Evaluation details and data are described in the appendix.

\subsection{Main results}
\label{sec:exp-main}

\begin{table}[t]
\centering
\scriptsize
\setlength{\tabcolsep}{2pt}
\renewcommand{\arraystretch}{1.02}
\caption{Question answering across general, temporal, and long-form benchmarks. RESUME uses at most $64$ I-frames, and each of the four predictive updates per I-frame produces one $8$-token readout. Prior numbers are as reported by CoPE-VideoLM (cited in the table); a dash means that source does not report the benchmark. Video-MME is without subtitles. ActivityNet-QA is scored by Claude Opus~4.8~\citep{anthropic2026claudeopus48} for RESUME and by a language model for the rows above.}
\label{tab:main}
\begin{tabular}{@{}c@{\hspace{6pt}}c@{\hspace{6pt}}c@{}}
\begin{tabular}[t]{@{}lrrrr@{}}
\multicolumn{5}{@{}l}{\textbf{(a) General QA}} \\
\toprule
& PT & NQA & AQA & VMME \\
\midrule
\multicolumn{5}{@{}l}{\cellcolor{white}\textit{Proprietary}} \\
\rowcolor{gen1}
{\shortstack[l]{GPT-5\\[-1pt]\scriptsize\citep{openai2025gpt5}}} & -- & 86.3 & -- & 83.3 \\
\rowcolor{gen1}
{\shortstack[l]{Gemini 3 Pro\\[-1pt]\scriptsize\citep{google2025gemini3}}} & -- & 84.3 & -- & 88.6 \\
\rowcolor{gen1}
{\shortstack[l]{Gemini 2.5 Pro\\[-1pt]\scriptsize\citep{gemini25_2025}}} & -- & 85.3 & -- & 87.8 \\
\rowcolor{gen1}
{\shortstack[l]{Claude 4.5\\[-1pt]\scriptsize\citep{anthropic2025claude45}}} & -- & 79.2 & -- & 74.2 \\
\multicolumn{5}{@{}l}{\cellcolor{white}\textit{Open-source}} \\
\rowcolor{gen2}
{\shortstack[l]{VILA-40B\\[-1pt]\scriptsize\citep{lin2024vila}}} & 54.0 & 67.9 & 58.0 & 60.1 \\
\rowcolor{gen2}
{\shortstack[l]{IXC-2.5-7B\\[-1pt]\scriptsize\citep{zhang2024ixc25}}} & 34.4 & 71.0 & 52.8 & 55.8 \\
\rowcolor{gen2}
{\shortstack[l]{LLaVA-OV-7B\\[-1pt]\scriptsize\citep{li2024llavaonevision}}} & 57.1 & 79.4 & 56.6 & 58.2 \\
\rowcolor{gen2}
{\shortstack[l]{Oryx-7B\\[-1pt]\scriptsize\citep{liu2024oryx}}} & 68.6 & 81.9 & -- & 58.3 \\
\rowcolor{gen2}
{\shortstack[l]{LLaVA-Video-7B\\[-1pt]\scriptsize\citep{zhang2024videoinstructiontuningsynthetic}}} & 67.9 & 83.2 & 56.5 & 63.3 \\
\rowcolor{gen2}
{\shortstack[l]{CoPE-7B\\[-1pt]\scriptsize\citep{cope2026}}} & 70.3 & 82.1 & 60.3 & 61.9 \\
\rowcolor{gen3}
\textbf{RESUME} & 71.2 & 81.9 & 60.9 & 61.9 \\
\bottomrule
\end{tabular}
&
\begin{tabular}[t]{@{}lrrr@{}}
\multicolumn{4}{@{}l}{\textbf{(b) Temporal}} \\
\toprule
& TC & TOM & MVB \\
\midrule
\multicolumn{4}{@{}l}{\cellcolor{white}\textit{Proprietary}} \\
\rowcolor{tmp1}
GPT-5 & 80.4 & 53.0 & 74.1 \\
\rowcolor{tmp1}
Gemini 3 Pro & 82.8 & 48.3 & 70.4 \\
\rowcolor{tmp1}
Gemini 2.5 Pro & 81.9 & 48.6 & 70.6 \\
\rowcolor{tmp1}
Claude 4.5 & 72.8 & 39.6 & 62.1 \\
\multicolumn{4}{@{}l}{\cellcolor{white}\textit{Open-source}} \\
\rowcolor{tmp2}
IXC-2.5-7B & 67.1 & -- & 69.1 \\
\rowcolor{tmp2}
LLaVA-OV-7B & 64.8 & 25.5 & 56.7 \\
\rowcolor{tmp2}
{\shortstack[l]{VideoLLaMA2\\[-1pt]\scriptsize\citep{damonlpsg2024videollama2}}} & -- & 18.5 & 54.6 \\
\rowcolor{tmp2}
{\shortstack[l]{InternVL2-8B\\[-1pt]\scriptsize\citep{chen2024internvl2}}} & 65.3 & 21.7 & 65.8 \\
\rowcolor{tmp2}
{\shortstack[l]{VideoChat2-7B\\[-1pt]\scriptsize\citep{2023videochat}}} & 45.5 & -- & 51.1 \\
\rowcolor{tmp2}
LLaVA-Video-7B & 66.6 & 24.9 & 58.6 \\
\rowcolor{tmp2}
CoPE-7B & 68.9 & 28.3 & 61.9 \\
\rowcolor{tmp3}
\textbf{RESUME} & 69.4 & 30.0 & 62.5 \\
\bottomrule
\end{tabular}
&
\begin{tabular}[t]{@{}lrrr@{}}
\multicolumn{4}{@{}l}{\textbf{(c) Long-form}} \\
\toprule
& VTT & VMMU & LVB \\
\midrule
\multicolumn{4}{@{}l}{\cellcolor{white}\textit{Proprietary}} \\
\rowcolor{lng1}
GPT-5 & -- & -- & 68.8 \\
\rowcolor{lng1}
Gemini 3 Pro & -- & -- & 78.0 \\
\rowcolor{lng1}
Gemini 2.5 Pro & -- & -- & 78.4 \\
\rowcolor{lng1}
Claude 4.5 & -- & -- & 50.5 \\
\multicolumn{4}{@{}l}{\cellcolor{white}\textit{Open-source}} \\
\rowcolor{lng2}
{\shortstack[l]{LongVA-7B\\[-1pt]\scriptsize\citep{zhang2024longva}}} & -- & 23.9 & -- \\
\rowcolor{lng2}
LLaVA-OV-7B & 44.0 & 33.9 & 38.1 \\
\rowcolor{lng2}
InternVL2-8B & -- & 37.4 & -- \\
\rowcolor{lng2}
LLaVA-Video-7B & 41.8 & 36.1 & 44.2 \\
\rowcolor{lng2}
CoPE-7B & 45.5 & 38.2 & 46.4 \\
\rowcolor{lng3}
\textbf{RESUME} & 45.7 & 38.2 & 43.2 \\
\bottomrule
\end{tabular}
\end{tabular}
\end{table}

\noindent\textbf{Results.}
RESUME shows consistent gains on the temporal benchmarks that directly test motion, order, and change. Compared with LLaVA-Video-7B, it improves by $2.8$, $5.1$, and $3.9$ points on TempCompass, TOMATO, and MVBench, respectively. Compared with the codec-based CoPE-7B baseline, the corresponding gains are $0.5$, $1.7$, and $0.6$ points. The improvements across both RGB-frame and codec-based baselines indicate that the carried state contributes beyond simply increasing temporal coverage: it provides a representation that preserves and accumulates predictive changes before they reach the language model.

On general and long-form video QA the picture is mixed. PerceptionTest and ActivityNet-QA improve over both LLaVA-Video-7B and CoPE-7B; Video-TT and Video-MMMU match CoPE-7B. Video-MME is $1.4$ points below LLaVA-Video-7B and tied with CoPE-7B; LVBench is $3.2$ points below CoPE-7B. Those two cells track the Stage-2 mixture: we fine-tune only on LLaVA-Video-178K, without the academic QA split or the image-alignment data used for the released base, and Video-MME is sensitive to that composition. Taken together, the stateful codec representation is most effective when the benchmark requires reasoning over temporal evolution, at the same per-predictive-frame readout budget.

\subsection{Motivation study: the information hierarchy}
\label{sec:exp-hierarchy}

\begin{figure}[t]
\centering
\includegraphics[width=\textwidth]{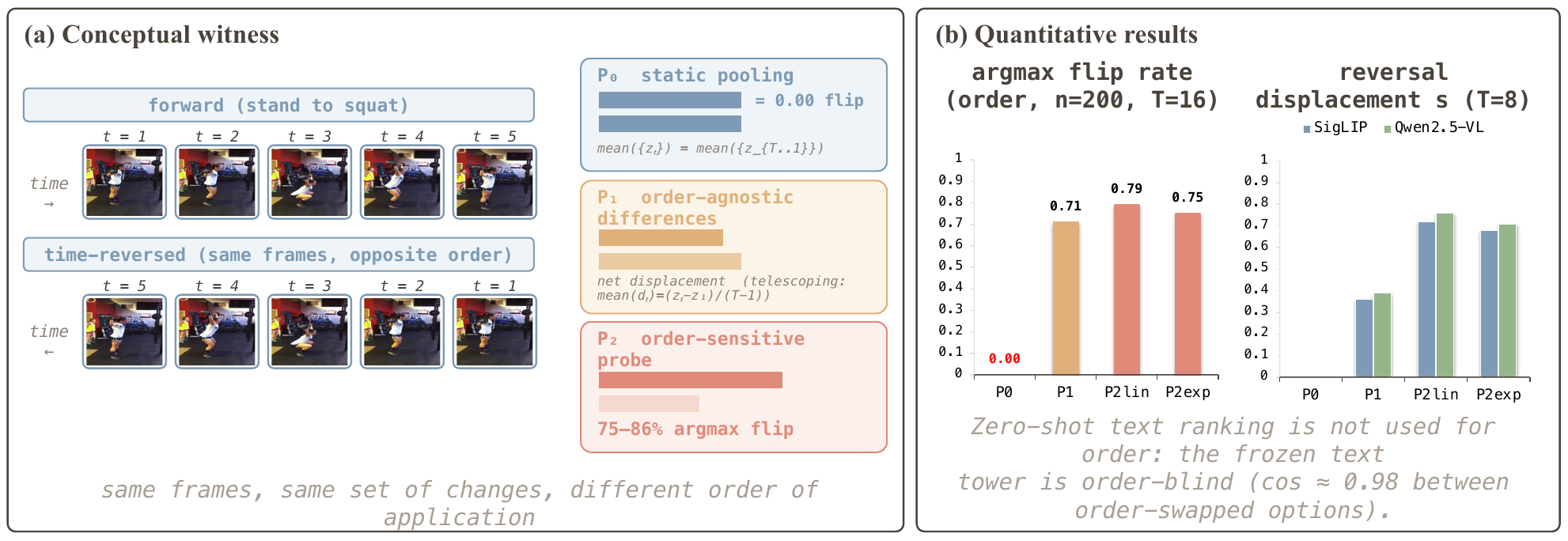}
\caption{The reversal axis separates static content, net change, and temporal order.
(a)~We compare frame-symmetric (P0), difference-symmetric (P1), and order-sensitive (P2) probes on a clip and its time reversal.
(b)~The probes are evaluated on order questions and on two frozen vision towers.}
\label{fig:hierarchy}
\end{figure}

The reversal experiment provides a direct separation witness for the information hierarchy. A clip and its time reversal contain exactly the same frames, so any representation that is symmetric over frames must produce the same score. As shown in Figure~\ref{fig:hierarchy}, static pooling (P0) is exactly reversal-invariant on both SigLIP and Qwen2.5-VL, with score displacement $0.0000$. The order-agnostic difference probe (P1) changes only through the net displacement between endpoints, as predicted by the telescoping identity, yielding an argmax flip rate of $0.68$--$0.71$. By contrast, the time-weighted order-sensitive probe (P2) flips on $0.75$--$0.86$ of clips. These results establish that order is a separate, measurable axis in frozen visual features rather than a by-product of static content or net change.

The same conclusion holds for the question-based probe. Order-swapped answer pairs are nearly indistinguishable in pooled text space ($\cos \approx 0.98$--$0.99$), whereas content-different pairs are substantially more separated ($0.58$--$0.72$). We therefore use QA ranking for the static and net-change buckets, and use the reversal axis to measure order sensitivity. Together, the two views motivate a representation that accumulates changes in sequence instead of reducing a clip to symmetric frame statistics. Full probe definitions and results are described in the appendix.

\subsection{Carried state between sampled frames}
\label{sec:exp-mech}

\begin{figure}[t]
\centering
\includegraphics[width=0.9\textwidth]{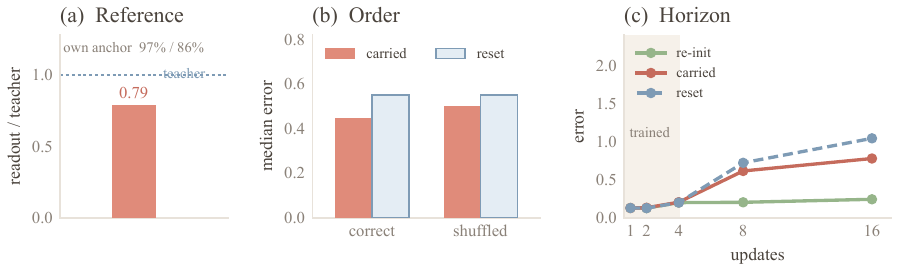}
\caption{The frozen transition carries state across predictive updates.
(a)~Anchor sensitivity, (b)~order sensitivity, and (c)~rollout behavior are compared with state controls.}
\label{fig:mech}
\end{figure}

To verify whether the frozen transition preserves anchor-dependent and ordered information, we conduct a study on $40$ videos from Video-MME~\citep{fu2024videomme} without language-model training; the results are summarized in Figure~\ref{fig:mech}. We compare readouts with frozen SigLIP features of the corresponding frames and report the median over videos.

To isolate anchor dependence, we apply the same updates to two different anchors. The resulting readouts move $0.79\times$ as far as the teacher features, while remaining closer to their respective anchors on $97\%$ and $86\%$ of videos (Figure~\ref{fig:mech}(a)). The transition therefore preserves anchor-dependent context while incorporating the current codec update.

Order sensitivity is examined by comparing correctly ordered and shuffled update sequences (Figure~\ref{fig:mech}(b)). The carried state produces lower error in the correct order and wins on $65\%$ of videos, whereas resetting to the anchor before every update removes order dependence and gives a median error of $0.55$. To probe behavior beyond the training horizon, we roll the transition forward for longer sequences (Figure~\ref{fig:mech}(c)). The carried state remains between the true-frame re-initialization lower bound and the memoryless control, and is more accurate than the latter on $75\%$ and $68\%$ of videos at horizons of $8$ and $16$ steps, respectively. These results show that the transition carries useful state beyond the immediate update and remains informative past the training horizon.

With a language model, we keep the base VideoLM's own sampled frames and interleave a readout of each predictive frame those samples skip. The frames, the vision tower, and the language model stay as in the base model; the readouts are how the changes between sampled frames enter it.

\subsection{Ablations}
\label{sec:exp-ablation}

To isolate the contribution of codec-driven state readouts, we construct paired evaluations under a matched I-frame budget. The base input contains at most $64$ I-frames, and each I-frame's four predictive updates each produce one $8$-token readout, giving $256$ readouts in total. For each choice of retained I-frames, $N_I\in\{8,16\}$, we compare two inputs built from exactly the same $N_I$ I-frames: one retains only the corresponding I-frame tokens, while the other augments those tokens with all $256$ state readouts. Since the I-frame content and sampling are identical within each pair, the performance difference isolates the contribution of the codec-driven state readouts.

\begin{table}[t]
\centering
\small
\caption{Ablation of the codec-driven state readouts under a matched I-frame budget. Each pair uses exactly the same $N_I$ I-frames; ``+\,readouts'' augments them with the $256$ state readouts. Accuracy (\%) on Video-MME (without subtitles) and LVBench.}
\label{tab:ablation}
\begin{tabular}{l >{\columncolor{gen1}}c >{\columncolor{gen2}}c >{\columncolor{lng1}}c >{\columncolor{lng2}}c}
\toprule
& \multicolumn{2}{c}{Video-MME} & \multicolumn{2}{c}{LVBench} \\
\cmidrule(lr){2-3}\cmidrule(lr){4-5}
$N_I$ & I-frames only & +\,readouts & I-frames only & +\,readouts \\
\midrule
$8$ & $53.3$ & $\mathbf{54.1}$\dbadge{0.8} & $36.7$ & $\mathbf{37.3}$\dbadge{0.6} \\
$16$ & $57.3$ & $\mathbf{58.6}$\dbadge{1.3} & $39.1$ & $\mathbf{40.0}$\dbadge{0.9} \\
\bottomrule
\end{tabular}
\end{table}

Table~\ref{tab:ablation} reports the comparison. Adding the state readouts improves accuracy over the I-frame-only input at both retained-I-frame budgets and on both benchmarks. At $8$ retained I-frames the readouts add $0.8$ points on Video-MME ($53.3\to54.1$) and $0.6$ on LVBench ($36.7\to37.3$); at $16$ I-frames they add $1.3$ points on Video-MME ($57.3\to58.6$) and $0.9$ on LVBench ($39.1\to40.0$). Because the two inputs of each pair are built from identical I-frames, these gains are attributable to the carried state alone rather than to additional I-frame coverage.

\section{Conclusion}
\label{sec:conclusion}

RESUME treats codec prediction as a representation-level state transition rather than a collection of independent predictive-frame token groups. By initializing a codec-conditioned latent state from each I-frame and updating it with fused motion-residual observations, the model preserves reference dependence and ordered change before the information reaches the language model. The experiments support this view: order is measurable in frozen visual features, the carried state responds to both anchors and update order, and the resulting readouts improve temporal reasoning over both RGB-frame and codec-based baselines at a matched per-predictive-frame budget. The state is carried only between successive I-frames: the next I-frame re-initializes it rather than updating it. We therefore do not claim a state held across a whole video; ordered change within a group of pictures is represented in the front-end, and aggregation across groups remains the language model's responsibility.

\bibliography{references}
\bibliographystyle{iclr2027_conference}

\clearpage
\appendix
\raggedbottom
\section{Training details}
\label{app:training}

The frozen vision encoder is SigLIP~\citep{siglip}. The latent state has $S{=}16$ slots of width $d_z{=}512$, and the readout uses $N{=}8$ tokens. Initialization and readout each use two pre-norm transformer layers. Motion vectors are patchified and embedded by a shared MLP; residuals are embedded by a truncated ResNet-18~\citep{He2015DeepRL} whose stride matches the motion grid.

Stage~1 trains only the state transition. The vision encoder stays frozen, and the objective is to regress each readout onto the frozen visual features of the corresponding frame. It unrolls $T{=}4$ ordered updates, and teacher features are pooled to the readout length. Optimization uses AdamW with peak learning rate $2\times10^{-4}$, weight decay $0.01$, and gradient clipping at $1.0$. The learning rate follows a cosine schedule: it warms up for $100$ steps and then decays over $1500$ steps to a quarter of the peak, $5\times10^{-5}$. Training uses $64$ GPUs at a global batch size of $16384$, in bfloat16. A readout is supervised at every step.

Stage~2 starts from LLaVA-Video-7B~\citep{zhang2024videoinstructiontuningsynthetic} and trains on LLaVA-Video-178K, with a context length of $32768$. The transition, the vision tower, and the original multimodal projector stay frozen; only the language model and a dedicated state projector are trained. The projector has the same shape as the multimodal projector but is independently initialized, because I-frame tokens represent absolute appearance and state readouts represent accumulated relative change. The objective is next-token prediction on the assistant text, and visual tokens are masked out of the loss. Optimization uses AdamW with peak learning rate $1\times10^{-5}$, weight decay $0$, and gradient clipping at $1.0$. The learning rate follows a cosine schedule with a floor of $1\times10^{-6}$ and a warmup ratio of $0.03$. The global batch size is $128$, and training runs for $10000$ steps. Each training video uses at most $64$ I-frames. Each I-frame is followed by four predictive updates, and one $8$-token readout is taken at each step.

\section{Training data construction}
\label{app:train-data}

We build the training data from the videos of LLaVA-Video-178K, pairing each video with its question--answer annotations at training time. Each video is partitioned into at most $64$ groups, adaptively to its duration so that longer videos yield more groups. Each group is initialized by an RGB anchor and consists of that anchor followed by four predictive updates; the predictive frames are sampled at a fixed stride of about one per second, shortened for very short groups, and we extract the motion vectors and residuals of each relative to the anchor, consistent with the definition of a codec P-frame. I-frames are resized to $384$ to match the vision-encoder input. At training time the visual content of each group---the I-frame through the vision encoder, the motion vectors and residuals through the state transition---is combined with the question--answer annotation of the corresponding video into a training example.

\section{Benchmarks}
\label{app:bench}

\paragraph{General video question answering.}
Video-MME~\citep{fu2024videomme} contains $900$ videos and $2{,}700$ multiple-choice questions, spanning six domains and three durations: short (under $2$ minutes), medium ($4$--$15$ minutes), and long ($30$--$60$ minutes). We report it without subtitles.

PerceptionTest~\citep{patraucean2023perception} (val) is a diagnostic suite over memory, abstraction, physics, and semantics. Its questions ask the model to describe, explain, predict, or consider a counterfactual, and are written so that a language prior is not enough to guess the answer.

NExT-QA~\citep{xiao2021next} is built from everyday videos and splits its five-way questions into causal (why or how an event happens), temporal (what happens before or after), and descriptive. Causal and temporal questions are the majority.

ActivityNet-QA~\citep{yu2019activityqa} asks open-ended questions about activities in web video, covering motion, objects, and time. Answers are short, and scoring is done by an external language-model judge.

\paragraph{Temporal and motion reasoning.}
TempCompass~\citep{liu2024tempcompass} pairs clips that share their static content and differ in one temporal aspect, so a single frame is not enough. The aspects are action, direction, speed, event order, and attribute change, and each is asked in more than one format, including multiple choice, yes/no, and caption matching.

TOMATO~\citep{shangguan2024tomato} contains $1{,}484$ human-annotated questions on $1{,}417$ videos, over action counting, direction, rotation, shape and trend, velocity and frequency, and visual cues. The clips are chosen so that one frame, or the frames in the wrong order, does not determine the answer.

MVBench~\citep{li2024mvbench} turns static image questions into twenty video tasks that a single frame cannot solve, from perception through to cognition. They include action sequence, movement direction, object interaction, scene transition, counting, and counterfactual inference.

\paragraph{Long-form video.}
LVBench~\citep{wang2024lvbench} uses videos of tens of minutes to several hours, with $1{,}549$ four-way questions whose number grows with duration. The questions cover long-range retrieval, tracking of entities and events, and reasoning across segments.

Video-TT~\citep{zhang2025videott} uses $1{,}000$ short videos. Each has one open question and four adversarial follow-ups, separating a failure of frame sampling from a failure to follow the visual or narrative content.

Video-MMMU~\citep{hu2025videommmu} uses $300$ instructional videos and $900$ questions across six disciplines. Each video is paired with questions at three stages of acquiring knowledge: perception, comprehension, and adaptation of what was learned to a new case.

\section{Evaluation data construction}
\label{app:eval-data}

We convert the raw videos of each evaluation benchmark into the same codec representation used at training time: an RGB anchor together with the motion vectors and residuals of its subsequent predictive frames. So that frame-indexed sampling is deterministic and reproducible, we first re-encode every video to a fixed structure of $30$ fps with I- and P-frames only and one I-frame every $8$ seconds. We then sample in the same units as training: each group consists of one RGB anchor followed by four predictive updates and does not cross an I-frame boundary. The number of groups adapts to video length, with at most $64$ groups per video. For each predictive update we read the encoder's motion vectors and resample them to a $24\times24$ macroblock grid (matching the $384$ input resolution), and take the residual as the difference between the true target frame and its motion-compensated prediction, consistent with the definition of a codec P-frame. Each video therefore yields up to $64$ I-frame anchors, and the four predictive updates of each group are read out as state readouts; the ablations in the main text either retain the I-frame tokens of only a subset of groups or drop the readouts entirely at evaluation time, without changing the underlying data.

\section{A group of pictures}
\label{app:codec}

\begin{figure}[H]
\centering
\includegraphics[width=0.98\textwidth]{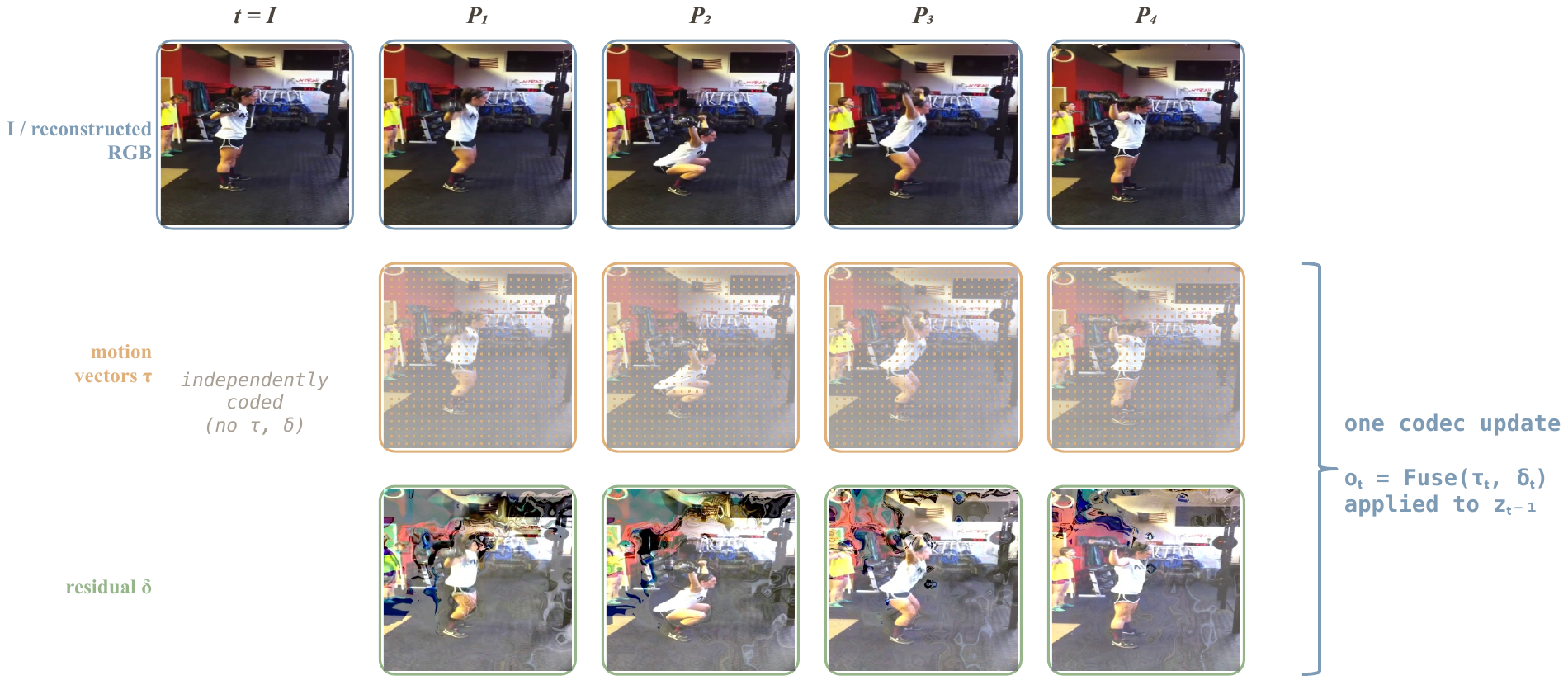}
\caption{Codec primitives inside one group of pictures.
The I-frame is an independently coded RGB image. Each predictive frame is stored as block-wise motion vectors $\tau$ (where existing content moves) and residuals $\delta$ (what motion compensation cannot explain). RESUME consumes $(\tau,\delta)$ as a single observation that updates a state initialized from the I-frame, rather than encoding each predictive frame as an independent token group.
Motion vectors are the bitstream quiver overlaid on a faded reconstruction of the same frame.}
\label{fig:codec}
\end{figure}

\section{Token budget at one frame per second}
\label{app:budget}

\begin{figure}[H]
\centering
\includegraphics[width=0.78\textwidth]{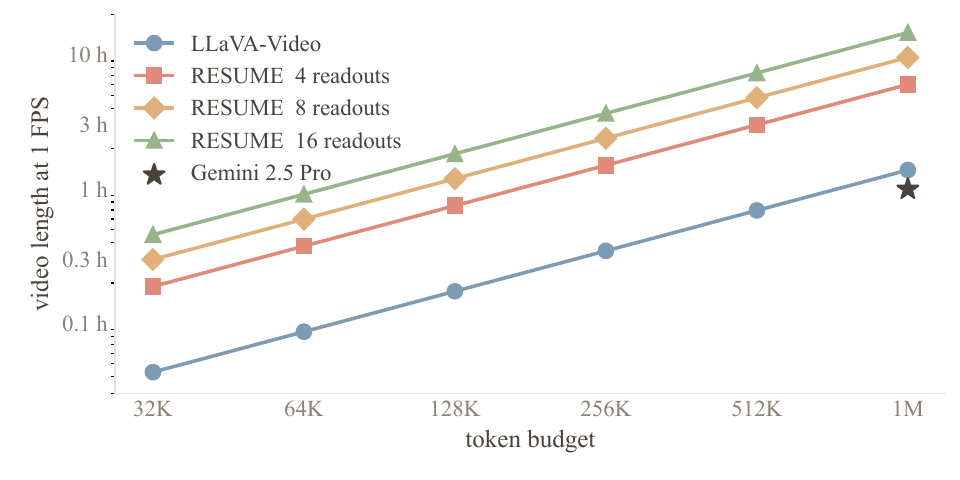}
\caption{Video length covered at one frame per second, against the visual-token budget. Markers distinguish dense frame encoding from three dense-readout settings ($4$, $8$, or $16$ eight-token readouts per I-frame). The star is the published Gemini~2.5 Pro point.}
\label{fig:budget}
\end{figure}

We compare visual-token budget with the video length it covers at one frame per second (Figure~\ref{fig:budget}). As a dense-readout upper bound, every predictive frame at one frame per second is read out as eight tokens, and we vary the number of readouts per I-frame. At one million tokens, $4$, $8$, and $16$ readouts per I-frame cover about $6$, $10$, and $15$ hours, respectively. Under the same budget, encoding every frame as LLaVA-Video does covers about $1.4$ hours, and the published working point of Gemini~2.5 Pro is about one hour~\citep{gemini25_2025}.

\section{Inference latency}
\label{app:latency}

Beyond question answering, we measure inference latency. Each sample spans about $64$ seconds at one frame per second. We generate $64$ text tokens with batch size $1$, greedy decoding, and a key--value cache.

Five inputs are compared. The first three cover the same span while reducing the I-frames that enter the language model and replacing the dropped anchors with additional groups of readouts: $32$ I-frames with $32$ readout groups, $16$ I-frames with $48$ readout groups, and $8$ I-frames with $56$ readout groups. The other two send no readouts: the same weights with $64$ I-frames only, and the original LLaVA-Video-7B with $64$ frames.

Time to first token is the prefill latency to the first text token. End-to-end latency is the time to generate $64$ tokens. After two warm-up runs we report the mean of five measurements.

\begin{table}[H]
\centering
\small
\caption{Inference latency for a $64$-second clip at one frame per second, generating $64$ text tokens. Time to first token (TTFT) is prefill; E2EL is the time to emit $64$ tokens.}
\label{tab:latency}
\begin{tabular}{lcc}
\toprule
Input & TTFT (s) & E2EL (s) \\
\midrule
\rowcolor{gen2}
$32$ I-frames + $32$ readout groups & $0.461$ & $1.795$ \\
\rowcolor{gen2}
$16$ I-frames + $48$ readout groups & $0.570$ & $1.895$ \\
\rowcolor{gen2}
$8$ I-frames + $56$ readout groups & $0.631$ & $1.954$ \\
\rowcolor{gen1}
$64$ I-frames (same weights) & $0.619$ & $1.938$ \\
\rowcolor{gen1}
$64$ frames, LLaVA-Video-7B & $0.686$ & $2.094$ \\
\bottomrule
\end{tabular}
\end{table}

Relative to $64$-frame LLaVA-Video-7B, all three readout configurations are faster: TTFT is lower by $0.225$, $0.116$, and $0.055$ seconds, and E2EL by $0.299$, $0.199$, and $0.140$ seconds. Relative to the same weights with $64$ I-frames only, $32{+}32$ lowers TTFT by $0.158$ seconds and $16{+}48$ by $0.049$ seconds. From $32{+}32$ to $16{+}48$ to $8{+}56$, fewer I-frames enter the language model but more groups are unrolled, and TTFT rises from $0.461$ to $0.570$ to $0.631$ seconds. At $8{+}56$ the first-token time is close to that of $64$ I-frames ($0.631$ versus $0.619$), so the extra state updates offset the saving from sending fewer I-frames. End-to-end latency follows the same order; decoding does not change it.

\section{Training-data scale and distribution}
\label{app:data-scale}

RESUME attaches the codec pathway to the released LLaVA-Video-7B and fine-tunes Stage~2 on LLaVA-Video-178K alone, without the three academic QA corpora or the LLaVA-OneVision image-alignment data that were used to build the released checkpoint. Our Stage-2 mixture is therefore both smaller and differently distributed than the data behind the base model.

Table~\ref{tab:data-scale} reproduces the incremental training stages of LLaVA-Video as reported in~\citep{zhang2024videoinstructiontuningsynthetic}. Video-MME shifts with the training mixture: the image-alignment corpus raises it ($63.2\to63.4$) and is absent from our mixture, while the academic QA split moves it the other way ($63.2\to61.9$). We include these rows only to note that Video-MME is sensitive to training-data scale and composition, and we do not attempt a controlled attribution. With this difference in mind, RESUME improves PerceptionTest ($+3.3$) and ActivityNet-QA ($+4.4$) over the base and lifts every temporal benchmark at a smaller and differently composed Stage-2 mixture.

\begin{table}[H]
\centering\scriptsize
\caption{Training-data scale and distribution. LLaVA-Video rows are as reported in~\citep{zhang2024videoinstructiontuningsynthetic} across incremental training stages; RESUME fine-tunes Stage~2 on LLaVA-Video-178K only. The three QA datasets are the training splits of PerceptionTest, NextQA, and ActivityNet-QA.}
\label{tab:data-scale}
\begin{tabular}{lcccc}
\toprule
\textbf{Training data} & \textbf{Total} & \textbf{NextQA} & \textbf{PerceptionTest} & \textbf{Video-MME} \\
\midrule
\multicolumn{5}{@{}l}{\cellcolor{white}\textit{LLaVA-Video~\citep{zhang2024videoinstructiontuningsynthetic}}} \\
\rowcolor{gen2}
LLaVA-Hound & $0.25$M & $64.4$ & $51.4$ & $54.1$ \\
\rowcolor{gen2}
\quad + LLaVA-Video-178K & $1.58$M & $80.1$ & $57.1$ & $63.2$ \\
\rowcolor{gen2}
\quad + 3 QA datasets & $1.64$M & $80.1$ & $69.0$ & $61.9$ \\
\rowcolor{gen2}
\quad + LLaVA-OV (images) & $2.74$M & $83.2$ & $67.9$ & $63.4$ \\
\rowcolor{gen2}
LLaVA-Video-178K (sampled) & $1.08$M & $73.2$ & $55.9$ & $59.6$ \\
\midrule
\rowcolor{gen3}
\textbf{RESUME} (Stage~2 on 178K) & $1.58$M & $81.9$ & $71.2$ & $61.9$ \\
\bottomrule
\end{tabular}
\end{table}

\section{Training-free probes of the information hierarchy}
\label{app:hierarchy}

A clip and its time reversal contain the same frames and the same average content; only the order in which changes were applied differs. Any representation that pools those frames symmetrically is invariant to the reversal by construction. Whether the discarded axis is empty in the frozen vision features that video language models actually use is a question that can be measured. We therefore build three parameter-free probes on the same frozen frame features, differing only in their symmetry over time. Let $z_1,\ldots,z_T$ be uniformly sampled frame features and $d_t=z_{t+1}-z_t$ the consecutive differences. The frame-symmetric probe takes the mean together with the two endpoints. Reversal only swaps the endpoints, the multiset of components is unchanged, and the score is identical. The difference-symmetric probe takes the mean, the standard deviation, and the maximum absolute value of the differences. The last two statistics are invariant to reversal. The mean obeys the telescoping identity $\mathrm{mean}(d)=(z_T-z_1)/(T-1)$ and only changes sign under reversal, so it carries net displacement rather than a path. The order-sensitive probe is a time-weighted sum of the differences, with weights given either by the time index, $\sum_t t\,d_t$, or by an exponential in that index, $\sum_t (0.8)^t d_t$. Because the weights are tied to time, reversal changes the sum. A probe is scored by the mean, over its components, of its cosine similarity with a candidate vector. Nothing is trained, and the probes have no parameters. We use $T\in\{8,16\}$, and features from two frozen towers that differ in both architecture and training objective: a contrastively trained SigLIP tower~\citep{siglip} and the vision tower of Qwen2.5-VL~\citep{Qwen2.5-VL}.

Questions are split by the information they require, with $200$ items in each bucket. The static bucket is drawn from the descriptive questions of NExT-QA~\citep{xiao2021next} and the action questions of TempCompass~\citep{liu2024tempcompass}; the net-change bucket from temporal questions about a final state and from attribute-change questions; the order bucket from questions about which event came first. Order cannot be witnessed by question-answering rank. Swapping ``$A$, and then $B$'' with ``$B$, and then $A$'' leaves the two sentences at cosine $0.98$--$0.99$ in pooled text space, against $0.58$--$0.72$ for pairs that differ in content. The zero-shot text criterion is itself nearly order-blind, so no video-side probe can rank such options by it. Question-answering rank is therefore used only on the static and net-change buckets. Evidence that order is separable comes from the reversal axis: the same clip played forward and backward, and whether the probe's decision changes. To compare the two towers without a shared text space, we also measure the displacement of the score vector against a fixed bank of random anchors.

On the $200$ order questions, the rate at which reversal flips the argmax matches these symmetries. At $T{=}8$ the frame-symmetric probe flips on none of the clips, the difference-symmetric probe on $0.68$, and the linear and exponential order-sensitive probes on $0.80$ and $0.86$. At $T{=}16$ the rates are $0.00$, $0.71$, $0.79$, and $0.75$. The zero of the frame-symmetric probe is an identity, not an estimate near zero. The flips of the difference-symmetric probe come from the sign change of its mean, at the magnitude the telescoping identity predicts. Across both weights and both lengths, the order-sensitive probe flips on $0.75$--$0.86$ of clips. On the same clips, the relative displacement of the anchor-bank score nearly coincides across towers. Under reversal at $T{=}16$, SigLIP moves by $0.0000$, $0.912$, $1.554$, and $1.430$ for the four probes in the order above; Qwen2.5-VL moves by $0.0000$, $0.892$, $1.566$, and $1.456$. At $T{=}8$ the frame-symmetric displacement is again exactly zero, the difference-symmetric displacement is $0.922$--$0.926$, and the order-sensitive displacements lie between $1.59$ and $1.71$.

Question-answering rank is readable only where the option text itself is separable. At $T{=}16$, scored against that text, the frame-symmetric probe reaches $0.685$ on the static bucket, while the three difference probes score $0.195$, $0.265$, and $0.290$, level with a random probe at $0.200$. Differences discard static appearance beyond the endpoints. On the net-change bucket every probe scores between $0.335$ and $0.365$, only slightly above the random probe at $0.290$, and no clean hierarchy appears. Ranking on the order bucket stays near that random floor, consistent with an order-blind text criterion, and is not offered as evidence about order.

Symmetric pooling therefore discards an axis that is not empty. In frozen visual features, a clip and its reverse are inseparable for the frame-symmetric probe, separable for an order-agnostic difference probe only through net displacement, and separable for a minimal order-sensitive probe. The same hierarchy recurs, at nearly the same values, on SigLIP and on Qwen2.5-VL, and is a property of frozen video representations rather than of one tower. Static appearance is carried by the endpoints and the mean. Order and content are different axes, not a refinement of one another.

Video language models encode sampled frames independently. When the number of frames is limited by the token budget, the changes between samples are dropped; pooling those frames symmetrically discards the order of the changes as well, by construction. This study shows that the discarded axis is non-empty in the feature spaces those models actually use, and that it can be read without training. What it supplies is the motivation for a stateful codec representation, not an ablation of one. A state updated in order applies the current observation to a reference it already carries, so order is preserved by construction and the readout stays in the space of visual features. The study trains no transition and compares no architecture. It only identifies the axis a representation has to keep.

\end{document}